\documentclass{article}

\usepackage{arxiv}
\usepackage[utf8]{inputenc} % allow utf-8 input
\usepackage[T1]{fontenc}    % use 8-bit T1 fonts
\usepackage{hyperref}       % hyperlinks
\usepackage{url}            % simple URL typesetting
\usepackage{booktabs}       % professional-quality tables
\usepackage{amsfonts}       % blackboard math symbols
\usepackage{nicefrac}       % compact symbols for 1/2, etc.
\usepackage{microtype}      % microtypography
\usepackage{graphicx}       % for figures
\usepackage{float}          % improved interface for floating objects
\usepackage{amsmath}
\graphicspath{ {./images/} } % path to your figure folder

\title{Project Jenkins:\\ Turning Monkey Neural Data into Robotic Arm Movement, and Back}

\author{
    Andrii Zahorodnii \\
    MIT \\
    \texttt{zaho@csail.mit.edu} \\
    \And
    Dima Yanovsky \\
    MIT \\
    \texttt{yanovsky@mit.edu} \\
}

\begin{document}
\maketitle

\begin{figure}[H]
    \centering
    \includegraphics[width=1.0\textwidth]{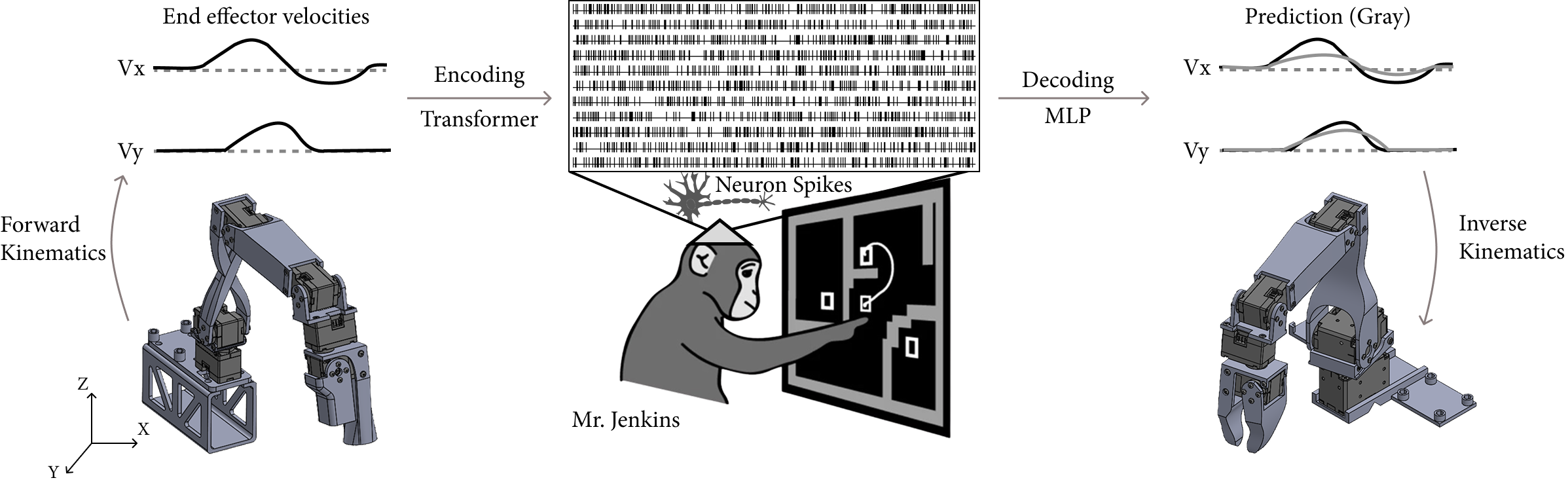}  % Replace with your actual figure file
    \caption{Project Jenkins. Leader arm velocities are computed via forward kinematics, then a transformer generates synthetic neural data. An MLP trained on real monkey neural data decodes it back into velocity space, commanding the follower arm’s movement. (Monkey diagram adapted from \cite{kaufman2014cortical}; robotic arm images from \cite{kochv1-1}).}
    \label{fig:jenkins_overview}
\end{figure}

%------------------------------------

\begin{abstract}

Project Jenkins explores how neural activity in the brain can be decoded into robotic movement and, conversely, how movement patterns can be used to generate synthetic neural data. Using real neural data recorded from motor and premotor cortex areas of a macaque monkey named Jenkins, we develop models for decoding (converting brain signals into robotic arm movements) and encoding (simulating brain activity corresponding to a given movement).

For the interface between the brain simulation and the physical world, we utilized Koch v1.1 leader and follower robotic arms. We developed an interactive web console that allows users to generate synthetic brain data from joystick movements in real time.
Our results are a step towards brain-controlled robotics, prosthetics, and enhancing normal motor function. By accurately modeling brain activity, we take a step toward flexible brain-computer interfaces that generalize beyond predefined movements. 

To support the research community, we provide open source tools for both synthetic data generation and neural decoding, fostering reproducibility and accelerating progress. The project is available at \url{https://www.808robots.com/projects/jenkins}.

\end{abstract}

\begin{figure}
    \centering
    \includegraphics[width=0.7\textwidth]{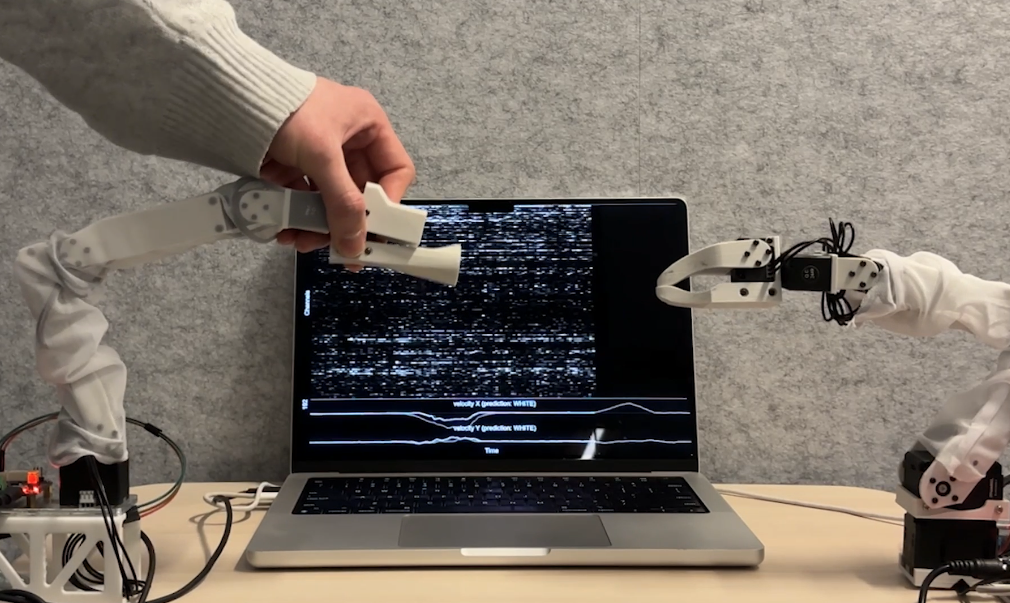} % Replace with your actual figure file
    \caption{Our approach in action. The experimenter moves the leader robotic arm, its velocity is recorded and transformed to synthetic neural data using the encoder model. Then, a decoder that was only trained on real neural data is decoding the neural data back into movement velocities, which are passed through inverse kinematics to the follower robotic arm. For the full video, please see the project’s webpage.}
    \label{fig:approach_in_action}
\end{figure}

% Introduction Section
%------------------------------------

\section{Introduction}
\label{sec:intro}
Synthetic neural data generation and neuroprosthetic devices are active areas of research, sparked by advances in neuroscience and robotics \cite{zhang2023noirneuralsignaloperated, burrow1997cortical, Afshar, Marrero2024}. These fields have significant implications for brain-computer interfaces, rehabilitation, and simulation of brain dynamics for downstream tasks or gaining new understanding of the underlying neural mechanisms.

In this project, which we call ``Project Jenkins,'' we explore such decoding and encoding of neural data from a macaque monkey named Jenkins. We used a publicly available dataset \cite{churchland2024} containing neural firing patterns from Jenkins' motor and premotor cortical areas during a center-outreach task.

Generating synthetic neural activity enables researchers to test and refine decoding models without requiring continuous access to live neural recordings \cite{kapoor2024latentdiffusionneuralspiking, Nakajima2023.03.05.531237}, while neuroprosthetic advancements \cite{pandarinath2017, willett2021brain2text, willett2023speech, gupta2023neuroprosthetics, doi:10.1126/scirobotics.adf5758,10.3389/fnbot.2016.00009, Gilja2012, doi:10.1126/scirobotics.adf7360} rely on robust encoding techniques to translate brain signals into precise motor commands. 

Our aim was two-fold (Figure \ref{fig:jenkins_overview}, \ref{fig:approach_in_action}):
\begin{itemize}
    \item \textbf{Decoding}: Translate neural spiking data into predicted velocities for a robotic arm.
    \item \textbf{Encoding}: Generate synthetic neural activity corresponding to an intended robotic movement.
\end{itemize}

With this paper, we publish the developed open-source tools for both synthetic neural data generation and neural decoding, enabling researchers to replicate our methods and build upon them. Our full codebase and additional resources, including demonstration videos, can be found on the project’s website: \url{https://www.808robots.com/projects/jenkins}.

%------------------------------------
% Data Collection and Preprocessing
%------------------------------------
\section{Methods and Results}

\subsection{Data Acquisition}
\label{sec:data}

\begin{figure}
    \centering
    \includegraphics[width=0.3\textwidth]{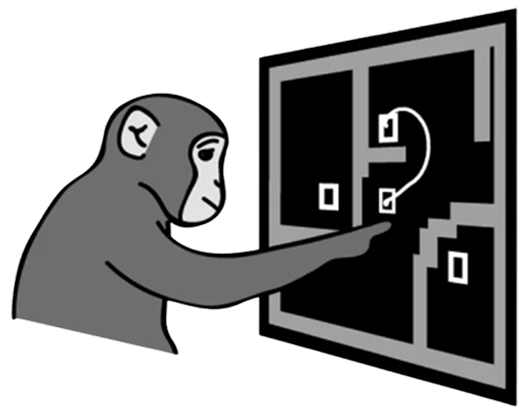}  % Replace with your actual figure file
    \caption{A typical monkey reaching task (adapted from \cite{kaufman2014cortical}). Jenkins starts with his hand at center, reaching out to one of 8 radial targets.}
    \label{fig:monkey_task}
\end{figure}

The neural data used in our project came from the primary motor cortex (M1) and caudal portion of the dorsal premotor cortex (PMd) areas of the brain of a rhesus macaque monkey, Jenkins. The dataset was published by Mark Churchland and others in 2021 \cite{churchland2024}. In short, Jenkins was trained on multiple reaching tasks, where the goal of the task is to press on dots that randomly light up on the screen in front of it (Figure \ref{fig:monkey_task}). Every time the monkey completes a trial successfully, he is rewarded with fruit juice. In our project, we used data from a center-outreach task, where the monkey always starts out with its hand in the middle of the screen, and the dots light up in one of 8 positions ($0^\circ, 45^\circ, 90^\circ, 135^\circ, 180^\circ, 225^\circ, 270^\circ,$ or $310^\circ$).

During these trials, neural spiking activity was measured with a multielectrode array capturing signals from $192$ distinct neurons. The final dataset contains thousands of trials across more than a dozen hours of recordings, accompanied by continuous measurements of Jenkins' hand position and velocity. In total, the spiking data was binned into discrete time intervals of $20\,\mathrm{ms}$.

All in all, there are more than a dozen hours of brain recordings together with the tracking of the monkey arm as it pressed dots on the screen thousands of times.

%------------------------------------
% Decoding: from Neural Data to Robot Movement
%------------------------------------

\subsection{Decoding: From Neural Data to Robot Movement}
\label{sec:decoding}

\begin{figure}
    \centering
    \includegraphics[width=0.8\textwidth]{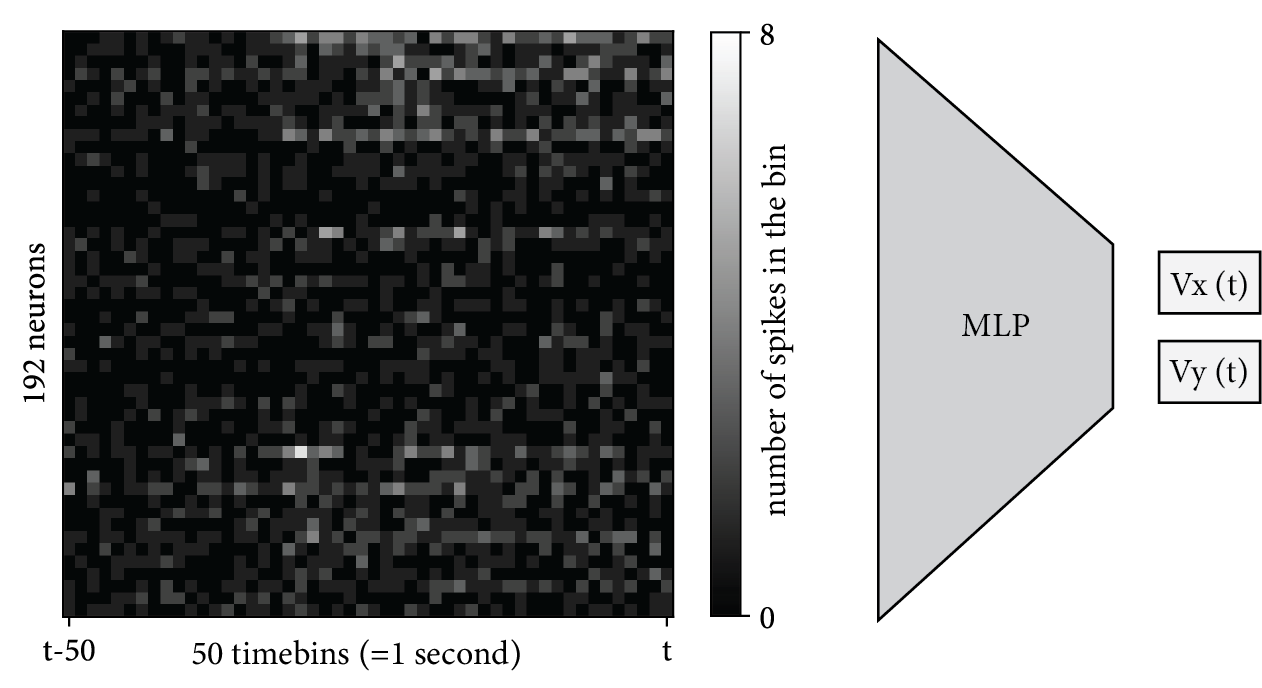}  % Replace with your actual figure file
    \caption{The decoding procedure.}
    \label{fig:decoding_procedure}
\end{figure}

Decoding is where we convert monkey brain data into the movements of the robotic arm. Whenever Jenkins wants to move his arm, neurons in his brain activate, calculating the trajectory of his planned movement, and this activation signal is sent through his spinal cord into the muscles in the monkey's arm. This sequence of events in time means that the movement of the arm depends on the recent history of neural data, and our model should take this fact into account. 

\textbf{Feature Construction.}
To predict the monkey's hand movements from brain recordings, we split time into bins of $20\,\mathrm{ms}$. For each bin, we recorded the total number of spikes from each of the $192$ neurons in that time bin (Figure \ref{fig:decoding_procedure}). This gives a vector $\mathbf{x}_t \in \mathbb{R}^{192}$ every $20\,\mathrm{ms}$. We also recorded the average $x$ and $y$ velocity of Jenkins' hand (from motion-capture data) during that same bin, forming a velocity vector $\mathbf{v}_t = (v_x, v_y)$.

\textbf{Neural Network Decoding Model.}
We employ a simple MLP (multilayer perceptron) with:
\begin{itemize}
    \item Two hidden layers (sizes $256$ and $128$).
    \item ReLU nonlinearities.
    \item Input window of $50$ bins (i.e., $1$ second of neural history).
\end{itemize}
Thus, at each time step $t$, we feed $\mathbf{x}_{t-49}, \dots, \mathbf{x}_t$ into the MLP to predict the current velocity $\mathbf{v}_t$. Despite its simplicity, this architecture performed effectively, achieving $R^2 \approx 0.9$ on a held-out test set.

\textbf{Driving the Robotic Arm.}
We implemented a \textit{follower} robotic arm using the Koch v1.1 design \cite{kochv1-1}. The arm has six servo motors arranged in a chain, and its end-effector position is determined by the servo angles. We decode velocity from the neural data, integrate it to obtain Cartesian coordinates $(X, Y)$, using the exponentially moving average (EMA), and then apply inverse kinematics (via the \texttt{ikpy} library, \cite{Manceron_IKPy}) to calculate servo angles for each motor. We defined a kinematic chain by specifying the servo motors' properties and the connecting link geometries. This chain configuration enables both forward kinematics (calculating end effector position from servo rotations) and inverse kinematics (determining required servo angles from desired coordinates). The EMA step is important to avoid the accumulation of errors over time. This loop yields continuous control of the robot in real time:
\[
\text{Neural Data} \;\Longrightarrow\; \hat{\mathbf{v}} \;\Longrightarrow\; (X, Y) \;\Longrightarrow\; \text{IK} \;\Longrightarrow\; \text{Servo Angles}.
\]

%------------------------------------
% Encoding: from Robot Movement to Neural Data
%------------------------------------

\subsection{Encoding: From Movement to Synthetic Neural Data}
\label{sec:encoding}

\begin{figure}
    \centering
    \includegraphics[width=0.9\textwidth]{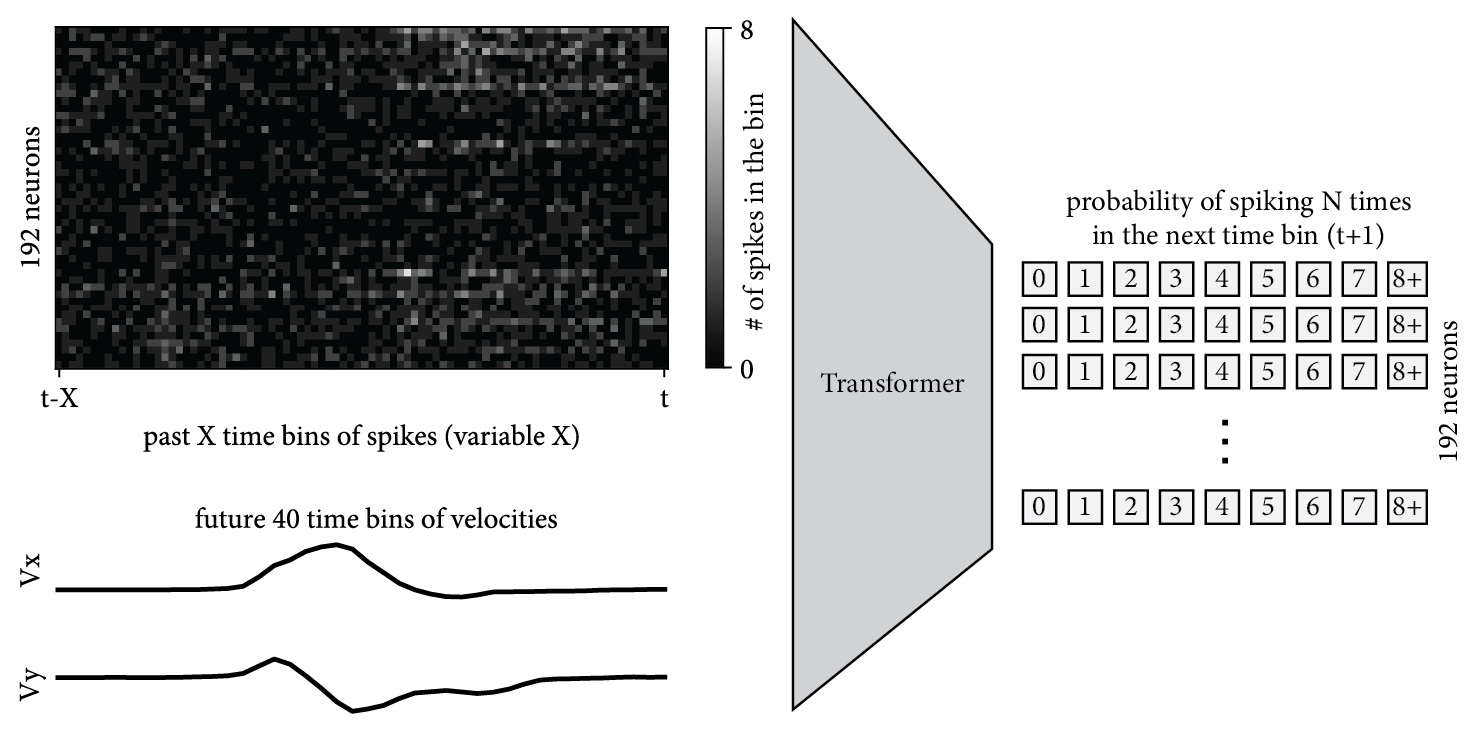}  % Replace with your actual figure file
    \caption{Generating synthetic neural data.}
    \label{fig:encoding_procedure}
\end{figure}

While the decoding problem was relatively straightforward, the encoding stage proved to be considerably more challenging. The encoding model is designed to \textit{generate} neural spiking patterns given a sequence of arm velocities (or positions).

\textbf{Closed-Loop Simulation Challenge.}
The key difficulty lies in closed-loop simulation: to continuously produce simulated neural data, the model needs to take its own past outputs as new inputs, to produce future predictions based on what it has already said before. So if a model is a little bit off every time, then the errors from its output get fed back into the model, producing even larger errors in the next output, and so on. 

Small errors can accumulate over time, destabilizing the generated signals. After several time bins, these errors would often blow up to large spiking rates or collapse to near-zero activity.
Success required fiddling with multiple different ways of formatting the inputs, finding the right architecture (transformer or LSTM) and training hyperparameters, as well as training for a long time (400 epochs).

\textbf{Transformer-based Encoder.}
Since the present movement of the arm depends on past neural data, it follows that the present neural data depends on the future (planned) movement of the arm. That is, to generate neural data, the model needs to know what the future arm movement will be. Accordingly, we process the data so that for every time bin, we input all of the past brain activity to the model, as well as the future arm movement velocities (Figure \ref{fig:encoding_procedure}). Specifically, we provided:
\begin{itemize}
    \item \textbf{Past neural activity}: Binned spike counts from previous time steps.
    \item \textbf{Future arm movement}: A ``look-ahead'' window of $40$ bins ($800\,\mathrm{ms}$) of velocities.
\end{itemize}
 
The model is trained to predict how many times each neuron will spike in the current $20\,\mathrm{ms}$ time bin. We formulate this problem as a 9-way classification: either a neuron will spike 0 times (be quiet), or 1, 2, …, 7, or 8+ times. 

The model was trained to output a 9-category distribution for each neuron, representing the number of spikes: $\{0,1,\dots,7,8+\}$
We train our encoding model to output what it thinks are the probabilities of each neuron's number of spikes being in any of our defined categories. This discrete classification formulation, akin to next-token prediction in language models \cite{radford2018improving,kaplan2020scaling}, helped stabilize training.

\textbf{Training Procedure}
We trained for approximately $400$ epochs, carefully tuning hyperparameters (learning rate, batch size, dropout) to avoid divergence. We found best results by setting learning rate to $0.0005$ and no weight decay ($\lambda_{wd}=0$). The training took under 4 hours on an off-the-shelf GPU with 12GB of RAM. We experimented with an LSTM-based approach \cite{hochreiter1997lstm} but eventually found a transformer architecture to be more robust. 

\subsection{Closing the Loop: Robot Movement to Neural Data and Back to Movement}
To accurately record the robotic arm's movement, we assembled a Koch v1.1 leader arm, sampling its $(x, y, z)$ coordinates at a frequency of 50 Hz. Since our study focuses exclusively on two-dimensional data, we discarded the $z$-coordinate. We then differentiated the $(x, y)$ positions to obtain the velocities $(V_x, V_y)$, which serve as input for our encoding model. 

After the model generates neural data corresponding to the movement inputs, we apply the decoding procedure to transform this synthetic neural data back into velocity components $(V_x, V_y)$. To reconstruct reliable spatial coordinates $(x, y)$ from these velocities, we employ an exponential decay filter $(\lambda_{\text{decay}} = 0.95)$, which mitigates the compounding of integration errors over time. 

Finally, the filtered positional data is passed to the Koch follower arm, where inverse kinematics algorithms compute the necessary servo rotations, enabling the follower arm to accurately replicate the original movements. The resulting system can be seen in Figure \ref{fig:approach_in_action} and on the project website.

%------------------------------------
% Conclusion
%------------------------------------

\subsection{Interactive Interface}
To enable users without direct robotic hardware experience to interactively generate neural data from movement, we developed a web application that allows joystick manipulation. This app records the velocities from joystick input and processes them through our transformer model to produce synthetic neural data, visualizing the output in real-time directly in the browser. Notably, the transformer model runs entirely within the user's browser rather than on a remote server. To achieve this, we converted our PyTorch model into a browser-compatible .onnx format using ONNX Runtime \cite{onnxruntime}, ensuring efficient local execution.

\section{Conclusion and Future Directions}
\label{sec:conclusion}
Project Jenkins demonstrates the feasibility of translating real neural activity into robotic arm movement and generating synthetic neural data to accompany or predict such movement. While our decoding model was successful and robust, the encoding model required more intricate architectures (transformers) and careful training to produce stable spike patterns over time. 

In practice, these methods can be extended to broader applications, such as human BCIs, prosthetics, and motor rehabilitation. Although our data was limited to eight primary reaching directions, preliminary experiments suggest generalization to more complex trajectories (e.g., drawing circles). Future work will focus on:
\begin{itemize}
    \item Testing on extended movement repertoires and more complex tasks.
    \item Improving the robustness of the encoding model in closed-loop simulation.
    \item Exploring real-time human-interface prototypes that adapt these neural decoders.
\end{itemize}
For more details, code, and demonstration videos, please visit the project webpage:
\url{https://www.808robots.com/projects/jenkins}.

\bibliographystyle{plain}
\bibliography{references}

\end{document}